\documentclass{article}

\PassOptionsToPackage{numbers, compress}{natbib}

\usepackage[preprint]{neurips_2025}

\usepackage[utf8]{inputenc} 
\usepackage[T1]{fontenc}    
\usepackage{hyperref}       
\usepackage{url}            
\usepackage{booktabs}       
\usepackage{amsfonts}       
\usepackage{nicefrac}       
\usepackage{microtype}      
\usepackage{xcolor}         

\usepackage{graphicx}
\usepackage{amsmath}
\usepackage{amssymb}
\usepackage{mathtools}
\usepackage{amsthm}
\usepackage{multirow}
\usepackage{makecell}
\usepackage{inconsolata}
\usepackage[english]{babel}
\usepackage{booktabs}

\PassOptionsToPackage{fnumbers,compress}{natbib}
\title{Information Gain-Guided Causal Intervention for Autonomous Debiasing Large Language Models}

\author{
\textbf{Zhouhao Sun}$^\spadesuit$\;\; \textbf{Xiao Ding}$^\spadesuit$\thanks{Corresponding Authors}\;\;\; \textbf{Li Du}$^\blacklozenge$\;\; \textbf{Yunpeng Xu}$^\spadesuit$\\ \textbf{Yixuan Ma}$^\spadesuit$\;\; \textbf{Yang Zhao}$^\spadesuit$\;\; \textbf{Bing Qin}$^\spadesuit$\;\; \textbf{Ting Liu}$^\spadesuit$ \\ $^\spadesuit$\footnotesize Research Center for Social Computing and Interactive Robotics\\\footnotesize Harbin Institute of Technology, Harbin, China \\
$^\blacklozenge$\footnotesize Beijing Academy of Artificial Intelligence, Beijing, China \\
\footnotesize\texttt{\{zhsun, xding, ypxu, yxma, yzhao, bqin, tliu\}@ir.hit.edu.cn}\\
\footnotesize\texttt{duli@baai.ac.cn}\\
}

\begin{document}

\maketitle

\begin{abstract}
Despite significant progress, recent studies indicate that current large language models (LLMs) may still capture dataset biases and utilize them during inference, leading to the poor generalizability of LLMs. However, due to the diversity of dataset biases and the insufficient nature of bias suppression based on in-context learning, the effectiveness of previous prior knowledge-based debiasing methods and in-context learning based automatic debiasing methods is limited. To address these challenges, we explore the combination of causal mechanisms with information theory and propose an information gain-guided causal intervention debiasing (ICD) framework. To eliminate biases within the instruction-tuning dataset, it is essential to ensure that these biases do not provide any additional information to predict the answers, i.e., the information gain of these biases for predicting the answers needs to be 0. Under this guidance, this framework utilizes a causal intervention-based data rewriting method to automatically and autonomously balance the distribution of instruction-tuning dataset for reducing the information gain. Subsequently, it employs a standard supervised fine-tuning process to train LLMs on the debiased dataset. Experimental results show that ICD can effectively debias LLM to improve its generalizability across different tasks.
\end{abstract}

\section{Introduction}
Large language models (LLMs), through the pre-training and supervised fine-tuning process, have demonstrated remarkable ability to follow human instructions to solve various tasks \citep{achiam2023gpt}, demonstrating immense potential in real-world applications.

\begin{figure}[t]
    \centering
    \includegraphics[width=1.01\linewidth]{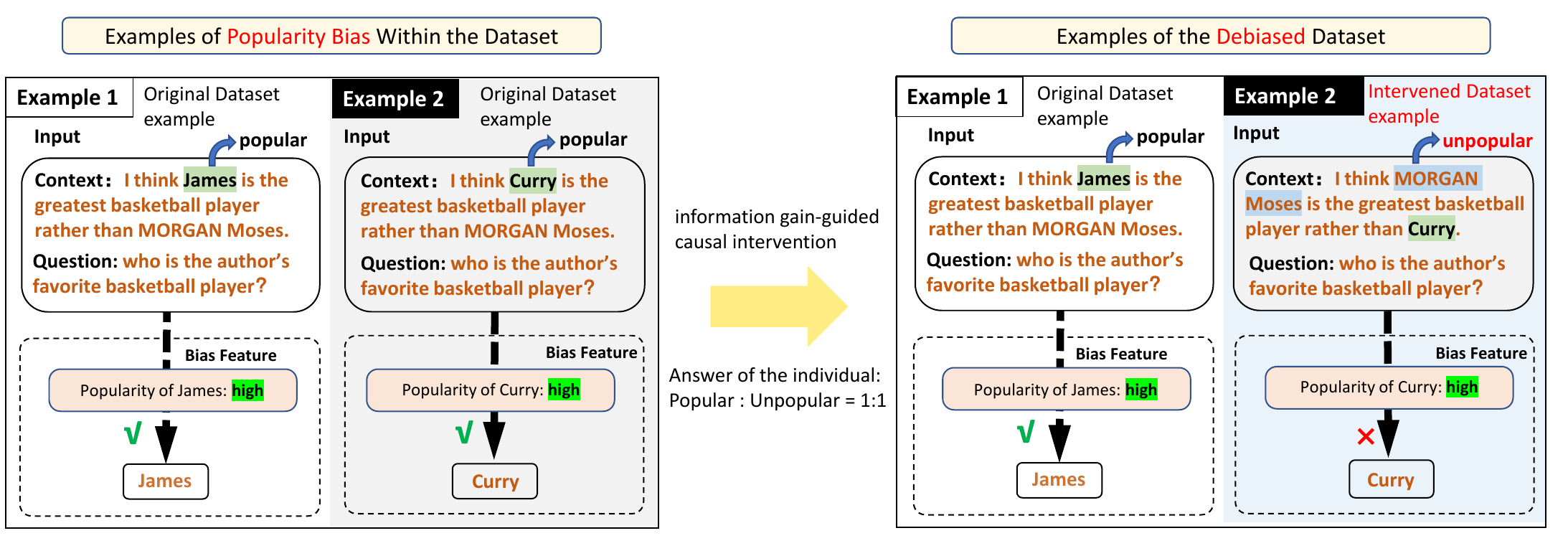}
    \caption{The examples in dataset before and after data debiasing. 
    By intervening example 2, an answer with a less-known individual is introduced (MORGAN Moses) so that LLMs cannot learn to make predictions solely relying on an individual's popularity. 
    }
    \label{fig:intro}
\end{figure}

Although achieving promising performance, LLMs would also inevitably learn \textbf{dataset bias} during the supervised fine-tuning process, such as negation bias and popularity bias \cite{schick2021self,mukherjee2021effect,navigli2023biases,klimashevskaia2024survey}. These dataset biases would lead to \emph{poor generalizability} of LLMs \citep{yang2024mitigating,cheng2024fairflow,2025RAZOR}. 
As shown in Figure~\ref{fig:intro}, 
the original dataset only contains samples where the answers are well-known figures, and it lacks samples featuring less-known individuals as answers.
Consequently, LLMs trained on such instruction-tuning corpus would also learn to utilize the biased feature `popularity of the individuals' for making predictions. However, when less-known people are generally the correct answer in some datasets, the performance of the LLM will significantly decline, indicating that dataset biases would reduce the generalizability of LLMs.


This problem highlights the necessity of debiasing LLMs. While contemporary debiasing methods are relatively effective, most of them require prior knowledge of dataset biases \citep{oba2023contextual,liu2023trustworthy,cheng2024fairflow}. However, dataset biases can vary significantly across different datasets and tasks \citep{poliak2018hypothesis,schuster2019towards,schick2021self}, making it  impractical to identify them one by one manually. Furthermore, a vast amount of biases still remain unrecognized in different datasets \citep{nie2020adversarial} and new biases are constantly being discovered. 

Hence, there is an urgent demand for methods to automatically debias LLMs without prior knowledge of dataset biases. However, previous automatic debiasing approaches \citep{sun2024causal} for LLMs rely on in-context learning (ICL) which can only suppress bias in LLMs instead of completely eliminating it in theory, as they mitigate bias through instructions without altering the model's intrinsic parameters. 

To address these issues, we start from the essential cause to debias the instruction-tuning dataset and propose an \textbf{I}nformation gain-guided \textbf{C}ausal intervention \textbf{D}ebiasing framework (ICD). 
To eliminate biased features within the instruction-tuning dataset, it is essential to ensure that these features do not provide any additional information to predict the answers, i.e., the information gain \citep{csiszar2015information} of these features for predicting the answers needs to be 0. Therefore, ICD initially employs an automatic bias detection method to identify biased features within the datasets. Subsequently, the information gain is utilized to check the relationships between biased features and answers. Following this, ICD utilize a causal intervention-based data rewriting method to intervene in the biased features. This process can modify the joint distribution between biased features and answers to eliminate their correlations, effectively reducing the information gain. Finally, LLMs are fine-tuned on the debiased dataset utilizing the supervised fine-tuning method. 

As illustrated in Figure~\ref{fig:intro}, by utilizing information gain-guided causal intervention method to revise example 2, the proportion of well-known individual (James) and less-known individual (MORGAN Moses) in the answer becomes the same. Consequently, the biased feature individual's popularity provide no information gain to predict the answers, indicating the success of the debiasing process.  

Experimental results show that our approach can improve the generalizability of LLMs on transfer and challenge datasets for four tasks. Furthermore, experiments on three general ability benchmarks demonstrate that IDB can improve the general ability of LLMs. We also note that datasets debiased by a certain LLM can also be utilized for another LLM to mitigate biases.

\section{Biased Features within the Instruction-tuning Datasets}
\label{gen_inst}

Text records and reflects human thoughts. However, due to inherent prejudice or preference of human beings, as well as potential annotation artifacts, various biases such as gender and popularity biases still broadly exist in various datasets \cite{steen2024bias,dai2024mitigate}. 

\begin{figure*}[t]
    \centering
    \includegraphics[width=1.0\linewidth]{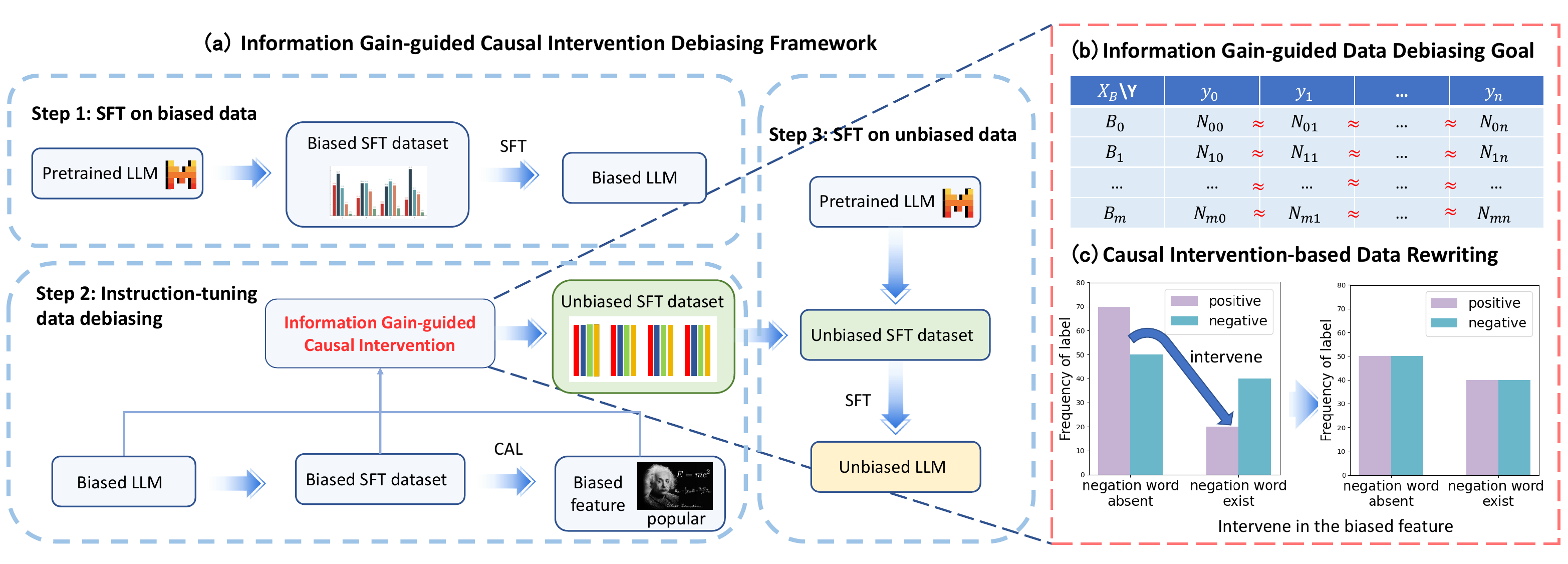}
    \caption{ (a) Three steps of Information Gain-Guided Causal Intervention  Debiasing Framework. (b) Illustration of the Information Gain-guided Data Debiasing Goal, i.e., the biased feature cannot contribute to correctly predicting the answers. (c) Illustration of the Causal Intervention-based Data Rewriting process.}
    \label{fig:method}
\end{figure*}

In instruction-tuning datasets, each piece of data consists of a question (i.e., instruction) and an answer. And each instruction-tuning dataset contains three types of features: The first category of features provides no assistance in accurately answering the questions within the dataset and is referred to as the non-predictive feature. The remaining features are collectively termed predictive features. Among these predictive features, causal features directly determine the answer and exhibit a causal relationship with it. The other predictive features, which do not have a direct causal relation to the answer, can be defined as predictive but non-causal features. For instance, in many datasets, the presence of a negation word such as `not' often appears when the sentiment of the sentence is negative rather than positive \citep{mukherjee2021effect}. However, the mere presence of the word `not' does not necessarily indicate that the sentiment is negative. In this case, the presence of the word "not" serves as a predictive but non-causal feature. In the debiasing scenario, the predictive but non-causal features are also referred to as biased features.

Among these categories of features, non-predictive features are generally not learned and utilized by powerful LLMs to answer questions, as they do not contribute to providing correct answers. Predictive features, which are useful for accurately answering questions within the dataset, are typically learned by LLMs. However, biased features lack a causal relationship with the answers. 
Considering that data distribution varies across different datasets, the predictive power of these biased features for answers may change. For example, if the sentiment of sentences in a dataset is generally positive when the negation word `not' appears, the performance of LLMs have learned to predict negative sentiment based on the negation word may decline.
Consequently, compared to learning causal features, the out-of-distribution generalization ability of LLMs may be decreased by learning biased features. 

\section{Method}
As illustrated in Figure~\ref{fig:method}~(a), ICD contains three main steps: (i) Fine-tune the pre-trained LLM on original biased instruction-tuning dataset; (ii) Biased feature identification and information gain-guided causal intervention for data debiasing. (iii) Fine-tune the pre-trained LLM on unbiased instruction-tuning dataset. 
Since (i) and (iii) are standard supervised fine-tuning processes, our focus will be on providing a detailed explanation on process (ii).

\subsection{Biased Feature Identification}
For brevity, we denote the large language models fine-tuned on the original biased instruction-tuning datasets as the biased LLM. This process aims at automatically identifying explainable biased features within the biased LLM without human prior knowledge. Specifically, we utilize the causal-guided active learning (CAL) method \citet{sun2024causal}, which is currently the only automatic method for identifying interpretable biased features. This method is applied independently to each task within the original instruction-tuning dataset to identify a series of biased features using the biased LLM. The identified biased features for each task can be seen in the Appendix~\ref{sec:biasfeature}.

\subsection{Information Gain-guided Data Debiasing Goal}
\label{sec:goal}
To debias the instruction-tuning dataset, we need to ensure that the identified biased features do not provide any additional information to predict the answers. That is, we need to make the information gain \citep{csiszar2015information} of these features for predicting the answers to be 0 under this dataset. Specifically, for any biased feature $B$:
\begin{align*}
\tag{1} IG(Y,B)=0.
\label{eq:1}
\end{align*} 
where $IG$ is the information gain, and $Y$ is the answer. From Eq.~\ref{eq:1}, we can know that:
\begin{align*}
\tag{2} H(Y|B)=H(Y).
\label{eq:2}
\end{align*} 
where H is the information entropy. By expressing $H(Y|B)$ and $H(Y)$ using probability formulas and extracting the public item, we have:
\begin{align*}
\tag{3} \sum_{b \in B} \sum_{y \in Y} P(b) P(y|b) (\log_2 P(y) - \log_2 P(y|b)) = 0.
\label{eq:3}
\end{align*}
The specific derivation process for Eq.~\ref{eq:3} can be found in Appendix~\ref{sec:proof}. 
By ensuring that $P(y)=P(y|b)$ for any values of y and b, Eq.~\ref{eq:3} can hold true constantly, that is to say: 
\begin{align*}
\tag{4} P(Y|B)=P(Y).
\label{eq:4}
\end{align*} 
Since there is no prior knowledge about $Y$, the probability distribution of $P(Y)$ should be uniform to avoid introducing label bias. Consequently, for any two values $y_i$ and $y_j$ of $Y$, we have:
\begin{align*}
\tag{5} P(y_i|B)=P(y_j|B).
\label{eq:5}
\end{align*} 
As we aims at debiasing the dataset, frequency is employed to replace probability. Consequently, given the dataset, for any value $b_k$ of $B$, we have
\begin{align*}
\tag{6} N(y_i,b_k)=N(y_j,b_k).
\label{eq:6}
\end{align*} 
where $N(y_i,b_k)$ denotes the number of instances whose answer and biased feature are $y_i$ and $b_k$, respectively. The Eq.~\ref{eq:6} formalizes the information gain-guided data debiasing goal. To the best of our knowledge, we are the first to establish strict criteria that an unbiased instruction-tuning dataset must satisfy. Note that even if the number $N$ exhibits slight disparities, LLMs are unlikely to learn these subtle correlations. Therefore, we relax the strict equality in Eq.~\ref{eq:6} to an approximate equality as shown in Figure~\ref{fig:method}~(b). 

For scenarios where the value space of the biased feature $B$ and answer $Y$ can be easily enumerated, such as the negation bias in the sentiment analysis task\footnote{A negation word either exists in the sentence or not, and the value space of the answer for sentiment analysis task is positive and negative.}, this data debiasing goal can directly guide the subsequent causal intervention process. 
However, for scenarios where the value space of the biased feature $B$ or the answer $Y$ is difficult to enumerate, we will utilize an automatic regularization method. The automatic regularization method can be divided into two aspects:
On the one hand, for the biased feature whose value space is a continuous space, it is normalized to a smaller, limited discrete space. For example, the value space of a person's popularity is originally a continuous space, but it can be directly normalized into "low popularity" and "high popularity" (this process can be automatically achieved by using LLMs). 
On the other hand, for answers whose value space is an extensive value space, automatic regularization can be executed based on the biased feature. Take the QA task and the individual's popularity as the biased feature as an example once more. The value space of the answer can potentially cover any person, so it has a vast value space. Hence, we can directly regularize the person according to the popularity (high or low). This process is also reasonable because when the biased feature under consideration is the individual's popularity, it does not matter who the people are appearing in the answer, but the popularity of the people matters. 
Other data debiasing goals that need to be regularized and the specific regularization process can be found in the Appendix~\ref{sec:adjustment}.

\subsection{Causal Intervention-based Data Rewriting}

The process of causal intervention \citep{pearl2009causality} involves modifying the original data distribution to produce a distribution that meets our expectations, which enables the realization of the information gain-guided data debiasing goal. 
Specifically, we first analyze the original distribution of the dataset and determine the intervention direction (e.g., the arrow in Figure~\ref{fig:method}~(c)). Subsequently, a causal intervention procedure is applied to a proportion of data by executing the operation $do(B=b_k)$, thereby modifying the data distribution to meet the data debiasing goal. This causal intervention process intervenes in the biased features, without changing the other semantics of the samples\footnote{For semantic that tightly tied to biased features, we also need to modify the semantics of the sample. For example, the individuals are tied to popularity, so we need to modify the figure when modifying popularity.}. 

For instance, as shown in Figure~\ref{fig:method}~(c), the label's frequency of the dataset is imbalanced when the negation word exists, suggesting that the existence of the negation word contributes to correctly predicting the answer of the dataset. 
In this example, $B$ represents the presence or absence of a negation word in a sentence, and we denote $b_k$ as the presence of the negation word. If we apply the operation $do(B=b_k)$ for a fixed number of samples that do not contain negation words and have a positive sentiment, they will be transformed into samples that include negation words while maintaining the same positive sentiment\footnote{For example, if the origin sample is `you will find yourself remembering this refreshing visit to a sunshine state', the operation $do(B=Yes)$ can be achieved by replacing the word `remembering' with `cannot forget'.}, so that the frequency of the label will be approximately balanced whenever the negation word exists, i.e., the data debiasing goal is satisfied. 

To perform the operation $do(B=b_k)$, we utilize the LLM itself to rewrite the data. Specifically, this causal intervention process employs the few-shot prompting method to rewrite samples within the dataset using the LLM itself. 
To ensure that the biased feature of the rewritten samples satisfies the intervention objectives $b_k$, we use the LLM itself or some metrics to check different biased features (specific details can be found in Appendix~\ref{sec:check}). If the requirements are met, the rewritten data is retained. Otherwise, the rewriting process is repeated until the intervenion objectives are met or the rewriting fails to meet the objectives three times.

Note that we also hope to minimize the amount of data to be rewritten, so the process of determining the intervention direction and the number of samples to be intervened can be regarded as a linear programming problem. Its objective function is to minimize the amount of rewritten data, and the constraint is that the distribution of the rewritten dataset should satisfy the information gain-guided data debiasing goal. By solving this linear programming problem, we can obtain the intervention direction and the number of samples to be intervened.

\section{Experimental Setup}
\subsection{Experimental Details}
In this work, we use Llama3.1-8B \cite{dubey2024llama} to debias the instruction-tuning dataset, and then fine-tune Llama3.1-8B on the debiased dataset for evaluating the ICD method. To examine the generalizablity of the debiased dataset, Gemma2-9B \cite{team2024gemma} is also utilized to fine-tune upon the debiased dataset.

Our instruction-tuning dataset consists of Lima \citep{zhou2024lima} and a portion of Flan 2021 \citep{wei2022finetuned} datasets. Due to the time limit, we separately choose 10,000 random samples from 5 sub-datasets of the Flan 2021 dataset, including MNLI \citep{williams2018broad}, QQP \citep{sharma2019natural}, SST2 \citep{socher2013recursive}, Squadv1 \citep{rajpurkar2016squad}, and TriviaQA \citep{joshi2017triviaqa}. These sub-datasets include four types of tasks: natural language inference (NLI), paraphrase identification (PI), sentiment analysis (SA), and question answer (QA). For NLI, SA and PI task, we ensure that the label is balanced when sampling. Since the task of the Lima dataset is multi-turn dialogue which is an open-ended task without standard answers, the biased features cannot be automatically induced by the CAL method \cite{sun2024causal}, and hence we apply the ICD method to the other four tasks.

We utilized few-shot prompt method for data rewriting and lora \citep{hu2022lora} for supervised fine-tuning (SFT) in our experiments. During the training process, we employed a batch size of 128 and a cosine learning rate schedule with an initial learning rate of 1e-4 for both models. We do not use early stopping, and train for three epochs instead. And each result is the average and the standard deviation of the scores across three different runs. Llama3.1-8B and Gemma2-9B are trained on 1 NVIDIA A100 80GB
PCIe GPU for 3 and 4 hours, respectively. The specific prompt for data rewriting can be seen in the supplemental material.

\subsection{Evaluation Tasks}
We examine the effectiveness of the information gain-guided causal intervention debiasing method by investigating whether it could debias LLMs to improve the zero-shot generalizability of LLMs. Specifically, we examine the generalizability of LLMs using transfer test sets and challenge test sets. 

\textbf{ID Test Sets}
We evaluate our debiasing framework on five in-domain (ID) datasets including MNLI, QQP, SST2, Squadv1, and TriviaQA. 

\textbf{Transfer Test Sets}
We evaluate the performance of LLMs in maintaining strong transferability across datasets for the same task, using SNLI \citep{bowman2015large}, MRPC \citep{dolan2005automatically}, IMDB \citep{maas2011learning}, and Squadv2 \citep{rajpurkar2018know} as transfer test sets for NLI, PI, SA, and QA tasks, respectively.

\textbf{Challenge Test Sets}
We assess the robustness of models against dataset bias using challenge test sets, specifically designed by manually eliminating the dataset biases from the dataset to challenge models. With the lack of corresponding challenge datasets for QA and SA task, we use HANS \cite{mccoy2019right} and PAWS \cite{zhang2019paws} as challenge test sets for NLI and PI task.

Furthermore, to evaluate the general ability of LLMs, we utilize the extensively used MMLU \citep{hendryckstest2021}, BBH \citep{suzgun2023challenging}, and TruthfulQA \citep{lin2022truthfulqa} datasets.
For BBH datasets, we use the exact match as the evaluation metric. For other datasets, we measured the accuracy to compare the results of different methods.

\subsection{Baseline Methods}
We compare the information gain-guided causal intervention debiasing method with two categories of training-based methods:

\begin{table}[tbp] 
\small
\caption{Comparison of ICD with baselines across two LLMs on ID and transfer test sets.}
\label{tab:main}
\centering
\setlength{\tabcolsep}{1mm}{
\begin{tabular}{ c|c|c|c|c|c|c|c c|c} 
\toprule 
\multicolumn{1}{c}{}&\multicolumn{2}{c}{NLI}&\multicolumn{2}{c}{PI}&\multicolumn{2}{c}{SA}&\multicolumn{3}{c}{QA}  \\ 
\cmidrule{2-10}
\multicolumn{1}{c}{Llama}&MNLI&SNLI &QQP&MRPC &SST2&IMDB &Squadv1&TriviaQA&Squadv2   \\ 
\hline 
\specialrule{0em}{1.5pt}{1.5pt}
\multicolumn{1}{c}{Vanilla} &84.3$\pm$0.2 &71.2$\pm$0.6 &\textbf{86.3}$\pm$0.1
 &75.3$\pm$0.4 &\textbf{95.4}$\pm$0.2 &95.4$\pm$0.1 &87.8$\pm$0.1 &65.6$\pm$0.1 &86.6$\pm$0.1     \\
\multicolumn{1}{c}{Razor}   &\textbf{84.5}$\pm$0.2&72.4$\pm$0.5 &85.5$\pm$0.1&75.0$\pm$0.5 &95.3$\pm$0.1&\textbf{95.5}$\pm$0.1 &87.8$\pm$0.1&65.6$\pm$0.1 &86.7$\pm$0.1    \\
\multicolumn{1}{c}{Ours}     &83.8$\pm$0.3&\textbf{75.2}$\pm$0.6 &85.7$\pm$0.1&\textbf{77.4}$\pm$0.2 &\textbf{95.4}$\pm$0.1&\textbf{95.5}$\pm$0.1 &\textbf{87.9}$\pm$0.1&\textbf{65.8}$\pm$0.1&\textbf{87.0}$\pm$0.1    \\
\hline 
\specialrule{0em}{1.5pt}{1.5pt}
\multicolumn{1}{c}{Gemma} &MNLI&SNLI &QQP&MRPC &SST2&IMDB &Squadv1&TriviaQA&Squadv2  \\ 
\hline 
\specialrule{0em}{1.5pt}{1.5pt}
\multicolumn{1}{c}{Vanilla} &76.2$\pm$0.2&63.6$\pm$0.6 &\textbf{87.0}$\pm$0.1&73.4$\pm$0.4 &\textbf{96.2}$\pm$0.2&95.4$\pm$0.1  &\textbf{85.8}$\pm$0.2 &68.9$\pm$0.1 &84.9$\pm$0.1    \\
\multicolumn{1}{c}{Razor}       &\textbf{76.3}$\pm$0.4&64.3$\pm$0.5 &86.5$\pm$0.1&72.5$\pm$0.4 &96.1$\pm$0.2&95.5$\pm$0.1  &85.6$\pm$0.2&68.9$\pm$0.2 &85.0$\pm$0.1        \\
\multicolumn{1}{c}{Ours}     &75.7$\pm$0.3&\textbf{65.8}$\pm$0.6 &86.5$\pm$0.1&\textbf{75.4}$\pm$0.5 &\textbf{96.2}$\pm$0.2&\textbf{95.7}$\pm$0.1  &85.5$\pm$0.1&\textbf{69.0}$\pm$0.2&\textbf{85.2}$\pm$0.2    \\
\bottomrule 
\end{tabular}
}
\end{table}

\vspace{+0.1cm}
\noindent\textbf{Vanilla Supervised Fine-tuning Baselines} 
We examine the performance of LLMs after vanilla supervised fine-tuning. 

\begin{table}[t]
\small
\begin{minipage}[l]{0.5\textwidth}
\caption{Comparison of ICD with baselines across two LLMs on ID and challenge test sets.}
\label{tab:challenge}
\centering
\setlength{\tabcolsep}{1mm}{
\begin{tabular}{ c|c|c|c|c} 
\toprule 
\multicolumn{1}{c}{}&\multicolumn{2}{c}{NLI}&\multicolumn{2}{c}{PI}\\ 
\cmidrule{2-5}
\multicolumn{1}{c}{Llama}&MNLI&HANS &QQP&PAWS   \\ 
\hline 
\specialrule{0em}{1.5pt}{1.5pt}
\multicolumn{1}{c}{Vanilla} &84.3$\pm$0.2&69.4$\pm$0.2 &\textbf{86.3}$\pm$0.1&62.2$\pm$0.2  \\
\multicolumn{1}{c}{Razor}       &\textbf{84.5}$\pm$0.2&70.1$\pm$0.3 &85.5$\pm$0.1&61.9$\pm$0.2 \\
\multicolumn{1}{c}{Ours}     &83.8$\pm$0.3&\textbf{72.3}$\pm$0.4 &85.7$\pm$0.1&\textbf{65.3}$\pm$0.2 \\
\hline 
\specialrule{0em}{1.5pt}{1.5pt}
\multicolumn{1}{c}{Gemma}&MNLI&HANS &QQP&PAWS   \\ 
\hline 
\specialrule{0em}{1.5pt}{1.5pt}
\multicolumn{1}{c}{Vanilla} &76.2$\pm$0.2&65.5$\pm$0.3 &\textbf{87.0}$\pm$0.1&57.1$\pm$0.2 \\
\multicolumn{1}{c}{Razor}       &\textbf{76.3}$\pm$0.4&65.7$\pm$0.3 &86.5$\pm$0.1&56.6$\pm$0.3 \\
\multicolumn{1}{c}{Ours}     &75.7$\pm$0.3&\textbf{67.0}$\pm$0.3 &86.5$\pm$0.1&\textbf{60.5}$\pm$0.4 \\
\bottomrule 
\end{tabular}
}
\end{minipage}
\quad
\small
\begin{minipage}[r]{0.46\textwidth}
\small
\caption{Comparison of ICD with baselines across two LLMs on the general ability benchmarks.}
\label{tab:general}
\centering
\setlength{\tabcolsep}{1.6mm}{
\begin{tabular}{ c|c|c|c} 
\toprule 
\multicolumn{1}{c}{Llama}&MMLU&BBH&TruthfulQA   \\ 
\hline 
\specialrule{0em}{1.5pt}{1.5pt}
\multicolumn{1}{c}{Vanilla} &60.3$\pm$0.3&41.5$\pm$0.4 &44.7$\pm$0.3           \\
\multicolumn{1}{c}{Razor}       &59.4$\pm$0.3&40.6$\pm$0.3 &45.6$\pm$0.4           \\
\multicolumn{1}{c}{Ours}     &\textbf{61.0}$\pm$0.3&\textbf{42.3}$\pm$0.3 &\textbf{52.8}$\pm$0.2  \\
\cmidrule{1-4}
\multicolumn{1}{c}{Gemma}&MMLU&BBH&TruthfulQA   \\ 
\hline 
\specialrule{0em}{1.5pt}{1.5pt}
\multicolumn{1}{c}{Vanilla} &68.3$\pm$0.1&45.3$\pm$0.1 &24.1$\pm$0.2          \\
\multicolumn{1}{c}{Razor}   &68.4$\pm$0.2&44.9$\pm$0.2 &23.5$\pm$0.3 \\
\multicolumn{1}{c}{Ours}    &\textbf{68.8}$\pm$0.2&\textbf{46.9}$\pm$0.3 &\textbf{27.3}$\pm$0.3 \\
\bottomrule 

\end{tabular}
}
\end{minipage}
\end{table}

\vspace{+0.1cm}
\noindent\textbf{Prior Knowledge-Based Debiasing Method}
Because there are no training-based automatic debiasing methods for LLMs. So we choose the latest prior knowledge-based debiasing method Razor \citep{2025RAZOR} and adapt it to the instruction tuning setting of LLMs. More details can be found in the Appendix~\ref{sec:RAZOR}.

\section{Experimental Results}
\subsection{Generalizability on Transfer Test Sets}
We list the experimental results of two LLMs on ID and transfer test sets in Table~\ref{tab:main}. From which we find that:

\begin{figure}[t]
\centering
\begin{minipage}[l]{0.48\textwidth}
    \includegraphics[width=1.0\linewidth]{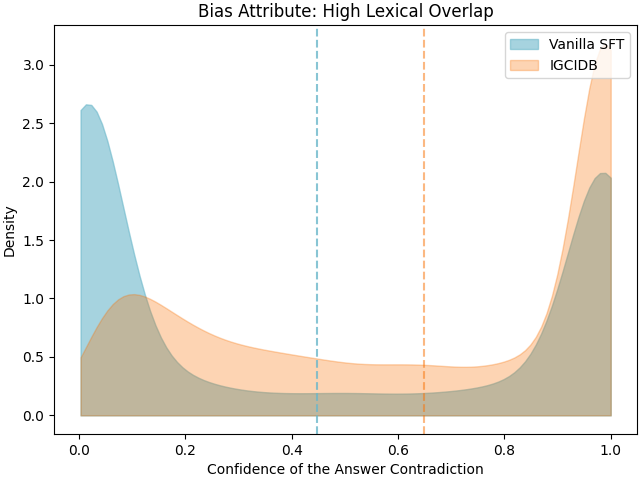}
    \caption{The density plot curve for the model confidence of answer contradiction when experimented with Llama3.1-8B.}
    \label{fig:density_llama3}
\end{minipage}
\quad
\begin{minipage}[r]{0.48\textwidth}
    \centering
    \includegraphics[width=1.0\linewidth]{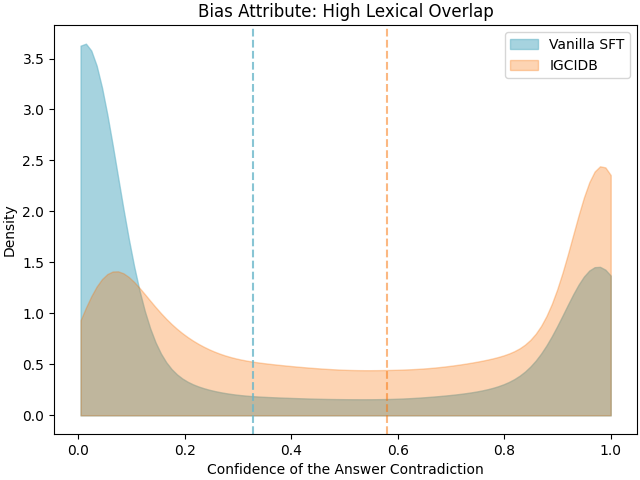}
    \caption{The density plot curve for the model confidence of answer contradiction when experimented with Gemma2-9B.}
    \label{fig:density_gemma2}
\end{minipage}
\end{figure}

\textbf{(1)} Compared to the vanilla supervised fine-tuning method, in general, the prior knowledge-based debiasing method Razor shows improved performance on the transfer test sets. This indicates that by utilizing prior knowledge to debias LLM, Razor can improve the generalizability of LLMs. 
However, it can be found that the performance of Razor on the paraphrase identification task is decreased. This may be due to the prior knowledge that the dataset bias originates solely from specific tokens in the input is not applicable for all tasks. This demonstrates the limitations of the prior knowledge-based debiasing method and further illustrates the importance of automatic debiasing methods for LLMs.

\textbf{(2)} Compared to the vanilla supervised fine-tuning method, ICD achieves consistent performance improvement on transfer test sets while maintaining performance on the in-domain datasets. 
This demonstrates that by leveraging the information gain-guided causal intervention, the biased features will not contribute to accurately predicting the answers for the instruction-tuning dataset. As a result, ICD can effectively debias LLMs by fine-tuning on the debiased dataset to improve the generalizability of LLMs.

\textbf{(3)} Compared with the prior knowledge-based debiasing method Razor, ICD shows relatively better performance on transfer test sets, which demonstrates the effectiveness of our method for debiasing LLMs. Additionally, compared to the vanilla supervised fine-tuning method, the performance of ICD increases on all transfer test sets, while the performance of Razor decreases on the MRPC dataset. On the one hand, the performance degradation of Razor shows the limitations of the prior knowledge-based debiasing method that the same prior knowledge may not be applicable to all tasks. On the other hand, the performance improvement of ICD on all transfer test sets further indicates the necessity of automatically identifying and eliminating dataset biases.

\textbf{(4)} In general, our method is effective for both Llama3.1-8B and Gemma2-9B. Moreover, even if fine-tuning Gemma2-9B on the dataset debiased using Llama3.1-8B, the effectiveness of debiasing to improve generalizability still remains. This demonstrates the generality of our approach in adapting to different LLMs. That is to say, once we debias a dataset using a certain LLM, we can directly use it to debias other LLMs. 

\subsection{Generalizability on Challenge Test Sets}
To investigate the generalizability of ICD on the manually debiased datasets, we examine our method on HANS and PAWS datasets. As shown in Table~\ref{tab:challenge}, compared to the prior knowledge-based debiasing method Razor and the vanilla supervised fine-tuning method, ICD shows better performance on both challenge test sets while maintaining performance on the in-domain datasets, which underscores the efficacy of our method for debiasing LLMs. 
Moreover, compared to the vanilla supervised fine-tuning method, the performance of ICD increases on both challenge test sets, while the performance of Razor decreases on PAWS. This shows the limitations of the prior knowledge-based debiasing method that the same prior knowledge may not be applicable to all tasks, and further demonstrates the necessity of automatically identifying and eliminating dataset biases. 

\begin{table}[t]
    \caption{Examples of case study. The baseline method tends to use the biased feature `individual's popularity' and to answer Scottish Parliament's member rather than an outside proposer in the first example. And it also utilizes the biased feature `high lexical overlap' to answer that the two sentences have thesame meaning in the third example.}
    \label{tab:case}
    \small
    \centering
    \begin{tabular}{lp{6.5cm}p{1.6cm}p{1.6cm}p{1.6cm}}
        \toprule
        \textbf{Task} & \textbf{Examples} & \textbf{Gold} & \textbf{Baseline} & \textbf{ICD} \\
        \midrule
        QA & Scottish Parliament: ... \textbf{a member of the Scottish Parliament} can introduce a member's bill as a private member; or a private bill can be submitted to Parliament by an outside proposer. ... Who may also submit private bills to the Parliament? & An outside proposer & Scottish Parliament's member \textcolor{red}{\texttimes} & An outside proposer \checkmark \\
        \addlinespace
        QA & Scottish Parliament: ... \textbf{a member of the Scottish Parliament} can introduce a member's bill as a private member; or a private bill can be submitted to Parliament by an outside proposer. ... Who can submit member's bill to the Parliament? & Scottish Parliament's member & Scottish Parliament's member \checkmark & Scottish Parliament's member \checkmark \\
        \addlinespace
        PI & Sentence 1: However, \textbf{Ann decides Peggy loses her visitation} as \textbf{the bed and breakfast is not safe}.
        Sentence 2: \textbf{Peggy decides} that \textbf{Ann loses her visitation} because \textbf{the bed and breakfast is not safe}.
        Do these sentences mean the same thing? & No & Yes \textcolor{red}{\texttimes} & No \checkmark \\
        \bottomrule
    \end{tabular}
\end{table}

\subsection{General Ability of LLMs}
Though effectiveness for debiasing LLMs, we hope that the general ability of LLMs will not be decreased compared to the vanilla supervised fine-tuning method. Therefore, we utilize MMLU, BBH, and TruthfulQA datasets to examine the general ability of different methods.

As shown in Table~\ref{tab:general}, the general ability of the ICD method is not decreased and even better than that of the vanilla supervised fine-tuning method in general, especially on TruthfulQA and BBH datasets. However, the performance of Razor is generally lower than the vanilla supervised fine-tuning method. This further indicates the effectiveness of our method.

\subsection{Analysis of Model Confidence}
To further investigate the effectiveness of the ICD method, we use the kernel density estimation to draw the distributions of model confidence on bias-conflicting samples. Specifically, we conduct experiments on the HANS dataset, where there is a high lexical overlap rate between the premise and hypothesis across all data. \citet{mccoy2019right} has demonstrated that for the NLI task, LLMs tend to predict entailment rather than contradiction when there is a high lexical overlap rate between the premise and hypothesis. Therefore, we consider the samples with label contradiction as the bias-conflicting samples. 

The result is illustrated in Figure~\ref{fig:density_llama3} and Figure~\ref{fig:density_gemma2}. It is observed that the vanilla supervised fine-tuning method shows high density on the low confidence of the answer contradiction, which demonstrates that LLMs tend to use the biased feature lexical overlap for predicting entailment after vanilla supervised fine-tuning. 
On the contrary, our method shows a much lower density on the low confidence of the answer contradiction compared to the vanilla supervised fine-tuning method. 
This is because, the information gain-guided causal intervention ensures that the biased feature `high lexical overlap' does not contribute to accurately predicting the answers. Consequently, LLMs will not exhibit obvious bias after fine-tuning on the debiased dataset.

\subsection{Case Study}
\label{sec:case}
To demonstrate the efficacy of the proposed method, we present a case study for QA ans PI task in Table~\ref{tab:case} (more case studies can be seen in the Appendix~\ref{sec:casestudy}). We compare our proposed method based on Llama3.1-8B with the baseline of the vanilla supervised fine-tuning method.

From Table~\ref{tab:case}, we can find that the baseline method tends to use the biased feature ``individual's popularity'' to answer Scottish Parliament’s member in both cases of the QA task. However, since the given context explicitly states ``a private bill can be submitted to Parliament by an outside proposer'', the correct answer is ``an outside proposer'' which is less popular than ``Scottish Parliament's member'' in the first example. Similarly, in the context of the PI task, the baseline method also tends to utilize the biased feature ``high lexical overlap'' to answer the meaning of the two sentences to be the same, failing to comparing the meaning of the two sentences. 

However, after applying the ICD method, our approach is capable of avoiding making inferences based on these biased features, thereby enabling the LLM to arrive at the correct answers.
Combining the case studies in Appendix~\ref{sec:casestudy}, it can be concluded that our method ICD, exhibits enhanced robustness compared to the baseline approach. The resilience of our method to conditions with different biased features suggests that it is well-suited for real-world environments where reliable LLMs are crucial.

\section{Related Work}
\textbf{LLMs Debiasing}
Previous works demonstrate that LLMs still suffer from biases such as negation bias \cite{mukherjee2021effect} and popularity bias \cite{klimashevskaia2024survey}.
To debias LLMs, a series of methods depend on researchers' prior knowledge to artificially detect the potential dataset biases, followed by debiasing through prompt-based regularization, instruction tuning, or preference optimization \cite{wang2023click,ganguli2023capacity,tanggeneralized2024GPO}. However, these methods are limited by the reliance on researchers' prior knowledge. Additionally, due to the diversity of dataset biases \cite{poliak2018hypothesis,schuster2019towards,schick2021self}, it is impractical to identify them one by one manually for all tasks and datasets. 

To address these issues, previous works proposed automatic debiasing methods. They train biased models to automatically extract biased features which characterizes the dataset biases \cite{utama2020towards,du2023towards,sanh2020learning,lyu2023feature}, and then some methods such as confidence regularization \citep{utama2020mind} are utilized for regularizing the main model. However, such methods are designed for discriminative models and are difficult to adapt to generative LLMs. To automatically debias generative LLMs, \citet{sun2024causal} proposed the causal-guided active learning method, which first combines the active learning and causal mechanisms to automatically induces explainable biased features, and then debias LLM utilizing in-context learning. However, this method is based on in-context learning so that it can only suppress bias in LLMs but cannot completely eliminate it in theory. Moreover, due to the limited instruction following ability of LLMs, it can only address limited numbers of biases. For example, it can only suppress two biases when utilizing Llama2-13b-Instruct \citep{touvron2023llama}.

In this paper, we propose an information gain-guided causal intervention debiasing framework for automatically debiasing generative LLMs. We borrow the idea from information theory and propose an information gain-guided data debiasing goal, followed by a causal intervention-based data rewriting method for debiasing instruction-tuning datasets. Based on the debiased dataset, the generalizability of LLMs can be improved using standard supervised fine-tuning.

\noindent\textbf{Causal Mechanism} Recently, a series of recent works have explored to debias the neural network models based on the causal mechanism. \citet{lv2022causality} utilized Structural Causal Model (SCM) to model auxiliary variables and exhibit promising performance in domain generalization. 
\citet{xu2023counterfactual} uses the causal effect estimation method to debias the language models for the fact verification task. \citet{lim2024identifying} applies the causal mediation analysis for debiasing NLU models. 
\citet{sun2024causal} explored the causal invariance mechanism to automatically identify biased samples and biased features. In this work, we combine information theory and the causal intervention \citep{pearl2009causality} method to automatically debias LLMs.

\section{Conclusions}
In this paper, we propose an information gain-guided causal intervention debiasing framework. According to the condition of an unbiased dataset that the biased feature should not provide any additional information for predicting the answers, we can derive the data debiasing goal using information theory. Then we utilize a causal intervention-based data rewriting method for achieving the data debiasing goal. Finally, a standard supervised fine-tuning process is used to debias LLMs. Experimental results show that our approach can effectively debias LLMs to enhance their generalizability in both transfer and challenge settings.

\bibliography{custom}
\bibliographystyle{plainnat}

\appendix

\section{Limitations}
Although our method can automatically debias LLMs, the bias identification process is based on the causal-guided active learning method, which cannot deal with tasks that do not have a standard answer such as writing essays.

\section{Proof of the information Gain-guided Data Debiasing Goal}
\label{sec:proof}
The entropy $H(Y)$ and the conditional entropy $H(Y|B)$ over all possible values of $Y$ and $B$ are defined as:
\begin{align*}
\tag{7} H(Y) = \sum_{y \in Y} P(y) \log_2 P(y),
\label{eq:7}
\end{align*} 
\begin{align*}
\tag{8} H(Y|B) = \sum_{b \in B} \sum_{y \in Y} P(y,b) \log_2 P(y|b),
\label{eq:8}
\end{align*} 
Substitute Eq.~\ref{eq:7} and Eq.~\ref{eq:8} into Eq.~\ref{eq:2}, we have
\begin{align*}
\tag{9} \sum_{y \in Y} P(y) \log_2 P(y) = \sum_{b \in B} \sum_{y \in Y} P(y,b) \log_2 P(y|b),
\label{eq:9}
\end{align*} 
By using probability formulas, we have
\begin{align*}
\tag{10} \sum_{b \in B} \sum_{y \in Y} P(b) P(y|b) \log_2 P(y) = \sum_{b \in B} \sum_{y \in Y} P(b) P(y|b) \log_2 P(y|b),
\label{eq:10}
\end{align*} 
Note that $\sum_{b \in B} \sum_{y \in Y} P(b) P(y|b)$ is a public item, so we have
\begin{align*}
\tag{11} \sum_{b \in B} \sum_{y \in Y} P(b) P(y|b) (\log_2 P(y) - \log_2 P(y|b)) = 0,
\label{eq:11}
\end{align*}
and finishes the proof.

\section{Data Debiasing Goal Based on Different Biased Features}
\label{sec:adjustment}
For each task, we choose the most prominent biased features (the first biased feature listed in Appendix.~\ref{sec:biasfeature}) identified by the CAL method for data debiasing. Specifically, for NLI and PI tasks, the most prominent biased feature is the lexical overlap between the two sentences. However, the lexical overlap rate can be any value between 0 and 1. To establish the data debiasing goal, the value space of this biased feature is divided into 3 classes: low lexical overelap (for lexical overelap rate which is less than 0.4), medium lexical overelap (between 0.4 and 0.6), low lexical overelap (bigger than 0.6). For these two tasks, the value space of the answer is easy to enumerate, so no modifications are necessary. For the SA task, the most prominent biased feature is whether the negation word exists in the sentence, so the data debiasing goal does not need to be regularized as illustrated in Sec.~\ref{sec:goal}. 
For the QA task, the most prominent biased feature is the popularity of a person or a location, and the regularization process has been shown in Sec.~\ref{sec:goal}.

\begin{table*}[t]
    \small
    \centering
    \begin{tabular}{lp{8cm}p{1cm}p{1cm}p{1cm}}
        \toprule
        \textbf{Task} & \textbf{Examples} & \textbf{Gold} & \textbf{Baseline} & \textbf{ICD} \\
        \midrule
        SA & If you're looking for a \textbf{not}-so-serious mob movie, with a female as the lead, you're in the right place. Pfieffer has acted much better than this. You can see she has matured beyond this picture. When I first picked this movie up, I expected Pfeiffer was poorly miscast, however, she plays her mob wife role to the hilt. \textbf{Not} a bad performance from Baldwin, either. If you \textbf{don't} pay attention to the hair, you might enjoy this movie. What is the sentiment of this review? & Positive & Negative \textcolor{red}{\texttimes} & Positive \checkmark \\
        \addlinespace
        NLI & Sentence 1: IDAs are special in that low-income savers receive matching funds from federal and state governments as well as private sector organizations as an incentive to save. Sentence 2: IDAs are special in that low-income savers receive differing funds from federal and state governments. Does the sentence 1 entail the sentence 2? & No & Yes \textcolor{red}{\texttimes} & No \checkmark \\
        \bottomrule
    \end{tabular}
    \caption{Examples of case study.}
    \label{tab:morecase}
\end{table*}

\section{Details of the biased feature checking process}
\label{sec:check}
To ensure that the biased feature of the rewritten samples satisfies the rewriting objectives, we employ different metrics to check these biased features. For lexical overlap bias, we calculate the lexical overlap rate between two input sentences. For negation word bias, we directly verify the presence of any predefined negation words in the inputs. For the popularity bias, we utilize the biased LLM to check if the individual or location is considered popular.



\section{Details of the Razor Baselines}
\label{sec:RAZOR}
The Razor method is based on the prior knowledge that the bias may originate from the specific token of the inputs. Based on this, it begins by defining a bias score to evaluate the degree of bias for each piece of data, followed by rewriting the data to achieve low bias scores for data debiasing. However, the calculation of the bias score requires to know the label space of the task, so that it cannot be applied to the QA and conversation task. As a result, when using it as a baseline, we apply this method to the MNLI, QQP, and SST2 datasets separately, and then merge them with the other unchanged instruction-tuning data to form a debiased instruction-tuning dataset for the subsequent supervised fine-tuning step.

\section{More Case Studies}
\label{sec:casestudy}
Here are case studies for the SA and NLI tasks. As illustrated in Table.~\ref{tab:morecase}, the first case contains negation words in the review, yet the sentiment is positive. The ICD method accurately predict its sentiment. However, the baseline method relies on the biased feature negation word to predict negative sentiment. This highlights the effectiveness of our method. For the example of the NLI task, there is a high lexical overlap rate between the two sentences, leading the baseline method to make incorrect predictions based on this biased feature. However, our method accurately predicts the correct answers for both samples. Combining the case studies in Sec.~\ref{sec:case}, it can be concluded that ICD exhibits improved robustness compared to the baseline approach.

\section{Identified biased features}
\label{sec:biasfeature}
Here we list three biased features identified by the CAL method for each task:

QA task: popularity of an individual or a location; social prominence of a person; nationality

NLI task: lexical overlap between the premise and hypothesis; semantic similarity between the premise and hypothesis; topic similarity between the premise and hypothesis

PI task: lexical overlap between the two sentences; syntactic similarity between the two sentences; keyword similarity between the two sentences

SA task: presence of negative words; complexity in language; the presence of certain words or phrases such as 'overcomes' or 'tragic'

\section{More Experimental Details}
For the evaluation of MNLI, QQP, MRPC, SST2, Squadv1, and Squadv2 datasets, 
we utilize the validation sets due to the unavailability of the test sets. For the other datasets, we use the test sets for evaluation.

\newpage
\section*{NeurIPS Paper Checklist}
\begin{enumerate}

\item {\bf Claims}
    \item[] Question: Do the main claims made in the abstract and introduction accurately reflect the paper's contributions and scope?
    \item[] Answer: \answerYes{} 
    \item[] Justification: The main claims can be supported by the experiments 

\item {\bf Limitations}
    \item[] Question: Does the paper discuss the limitations of the work performed by the authors?
    \item[] Answer: \answerYes{} 
    \item[] Justification: We provide the limitations of this work in Appendix A.

\item {\bf Theory assumptions and proofs}
    \item[] Question: For each theoretical result, does the paper provide the full set of assumptions and a complete (and correct) proof?
    \item[] Answer: \answerYes{} 
    \item[] Justification: We provide the full set of assumptions and a complete proof in Appendix B and Section 3.2.

    \item {\bf Experimental result reproducibility}
    \item[] Question: Does the paper fully disclose all the information needed to reproduce the main experimental results of the paper to the extent that it affects the main claims and/or conclusions of the paper (regardless of whether the code and data are provided or not)?
    \item[] Answer: \answerYes{} 
    \item[] Justification:  We provide the training and evaluation details for reproducibility in Section 4. We also submit the code and data in the supplemental material.

\item {\bf Open access to data and code}
    \item[] Question: Does the paper provide open access to the data and code, with sufficient instructions to faithfully reproduce the main experimental results, as described in supplemental material?
    \item[] Answer: \answerYes{} 
    \item[] Justification: We have provided the the data and code in the supplemental material.

\item {\bf Experimental setting/details}
    \item[] Question: Does the paper specify all the training and test details (e.g., data splits, hyperparameters, how they were chosen, type of optimizer, etc.) necessary to understand the results?
    \item[] Answer: \answerYes{} 
    \item[] Justification: We have provided the training and test details in Sec 4. Some other details such as the biased features and prompts for data rewriting are provided in appendix and supplemental material.

\item {\bf Experiment statistical significance}
    \item[] Question: Does the paper report error bars suitably and correctly defined or other appropriate information about the statistical significance of the experiments?
    \item[] Answer: \answerYes{} 
    \item[] Justification: We have reported the standard deviation across 3 runs in Table 1, 2 and 3.

\item {\bf Experiments compute resources}
    \item[] Question: For each experiment, does the paper provide sufficient information on the computer resources (type of compute workers, memory, time of execution) needed to reproduce the experiments?
    \item[] Answer: \answerYes{} 
    \item[] Justification: We have provided the compute resources and time of execution in Sec 4.
    
\item {\bf Code of ethics}
    \item[] Question: Does the research conducted in the paper conform, in every respect, with the NeurIPS Code of Ethics \url{https://neurips.cc/public/EthicsGuidelines}?
    \item[] Answer: \answerYes{} 
    \item[] Justification: Our research does not involve human subjects or participants. We also follow the licenses when using publicly available datasets.

\item {\bf Broader impacts}
    \item[] Question: Does the paper discuss both potential positive societal impacts and negative societal impacts of the work performed?
    \item[] Answer: \answerNA{} 
    \item[] Justification: We focus on the generalization abilities of LLMs, which is a foundational research and does not concentrate on specific application scenarios.
    
\item {\bf Safeguards}
    \item[] Question: Does the paper describe safeguards that have been put in place for responsible release of data or models that have a high risk for misuse (e.g., pretrained language models, image generators, or scraped datasets)?
    \item[] Answer: \answerNA{} 
    \item[] Justification: We use publicly available data and LLMs for experiments.

\item {\bf Licenses for existing assets}
    \item[] Question: Are the creators or original owners of assets (e.g., code, data, models), used in the paper, properly credited and are the license and terms of use explicitly mentioned and properly respected?
    \item[] Answer: \answerYes{} 
    \item[] Justification: We cite every resource (such as LLMs and benchmarks) used in this work in Section 4. We also follow the the license and terms of use of corresponding LLMs and data.

\item {\bf New assets}
    \item[] Question: Are new assets introduced in the paper well documented and is the documentation provided alongside the assets?
    \item[] Answer: \answerYes{} 
    \item[] Justification: We provide the code to create the debiased dataset and debiased LLMs in the supplemental material.

\item {\bf Crowdsourcing and research with human subjects}
    \item[] Question: For crowdsourcing experiments and research with human subjects, does the paper include the full text of instructions given to participants and screenshots, if applicable, as well as details about compensation (if any)? 
    \item[] Answer: \answerNA{} 
    \item[] Justification: This paper does not involve crowdsourcing nor research with human subjects.

\item {\bf Institutional review board (IRB) approvals or equivalent for research with human subjects}
    \item[] Question: Does the paper describe potential risks incurred by study participants, whether such risks were disclosed to the subjects, and whether Institutional Review Board (IRB) approvals (or an equivalent approval/review based on the requirements of your country or institution) were obtained?
    \item[] Answer: \answerNA{} 
    \item[] Justification: This paper does not involve crowdsourcing nor research with human subjects

\item {\bf Declaration of LLM usage}
    \item[] Question: Does the paper describe the usage of LLMs if it is an important, original, or non-standard component of the core methods in this research? Note that if the LLM is used only for writing, editing, or formatting purposes and does not impact the core methodology, scientific rigorousness, or originality of the research, declaration is not required.
    \item[] Answer: \answerNA{} 
    \item[] Justification: We use LLM for writing and editing.

\end{enumerate}

\end{document}